\documentclass[runningheads]{llncs}
\usepackage{graphicx}
\usepackage{amsmath,amssymb} 
\usepackage{color}
\usepackage[width=122mm,left=12mm,paperwidth=146mm,height=193mm,top=12mm,paperheight=217mm]{geometry}

\usepackage{times}
\usepackage{epsfig}
\usepackage{graphicx}
\usepackage{amsmath}
\usepackage{amssymb}
\usepackage{float}
\usepackage{enumerate}
\usepackage{xfrac}
\usepackage{url}

\begin{document}
\pagestyle{headings}
\mainmatter
\def\ECCV16SubNumber{1122}  

\title{The Conditional Lucas \& Kanade Algorithm} 

\author{Chen-Hsuan Lin \and Rui Zhu \and Simon Lucey}
\institute{The Robotics Institute, Carnegie Mellon University \\
{\tt\small \{chenhsul,rz1\}@andrew.cmu.edu, slucey@cs.cmu.edu}}

\maketitle

\newcommand{\fig}[1]{Figure \ref{Fig:#1}}
\newcommand{\eqn}[1]{Eqn. \ref{eqn:#1}}
%
\newcommand{\eucsqnorm}[1]{\left\|{#1}\right\|_2^2}
\newcommand{\frobnorm}[1]{\left\|{#1}\right\|_F^2}

\newcommand{\I}{\mathcal{I}}
\newcommand{\T}{\mathcal{T}}
\newcommand{\p}{\mathbf{p}}
\newcommand{\0}{\mathbf{0}}
\newcommand{\deltap}{\Delta \p}
\newcommand{\Ip}{\I(\p)}
\newcommand{\TO}{\T(\0)}
\newcommand{\y}{\mathbf{y}}
\newcommand{\x}{\mathbf{x}}
\newcommand{\dTOdp}{\frac{\partial \T(\0)}{\partial \p}}
\newcommand{\dxdp}{\frac{\partial \mathcal{W}(\x; \0)}{\partial \p^{\top} }}
\newcommand{\R}{\mathbf{R}}
\newcommand{\G}{\mathcal{G}}
\newcommand{\g}{\mathbf{g}}
\newcommand{\gj}{g_{j}}
\newcommand{\deltag}{\Delta \g}
\newcommand{\Gg}{\mathcal{G}(\g)}
\newcommand{\ghat}{\mathbf{\hat{g}}}
\newcommand{\Rg}{\mathcal{R}(\g)}
\newcommand{\dRgdg}{\frac{\partial \Rg}{\partial \g}}
\newcommand{\dRgdgj}{\frac{\partial \Rg}{\partial \gj}}
\newcommand{\Hg}{\mathcal{H}(\g)}
\newcommand{\dHgdgj}{\frac{\partial \Hg}{\partial \gj}}
\newcommand{\dinvHgdgj}{\frac{\partial (\Hg)^{-1}}{\partial \gj}}
\newcommand{\indMtrxj}{\mathbf{\Lambda}_j}
\newcommand{\eye}{\mathbf{I}}
\newcommand{\Real}{\mathbb{R}}
\newcommand{\lbar}{\left\|}
\newcommand{\rbar}{\right\|}
\newcommand{\dTOdpinline}{\sfrac{\partial \T(\0)}{\partial \p}}
\newcommand{\dxdpinline}{\sfrac{\partial \x}{\partial \p}}
\newcommand{\deltapitrain}{\delta \p_i}
\newcommand{\deltapItrain}{\delta \p_1}
\newcommand{\deltapNtrain}{\delta \p_N}
\newcommand{\deltaxitrain}{\delta \x_i}
\newcommand{\z}{\mathbf{z}}

\renewcommand{\S}{\mathcal{S}}
\newcommand{\W}{\mathbf{W}}
\newcommand{\X}{\mathbf{X}}
\newcommand{\Y}{\mathbf{Y}}
%
%
\newcommand{\diag}{\mbox{diag}}
\newcommand{\kron}{\otimes}
\newcommand{\N}{\mathcal{N}}
\renewcommand{\vec}{\mbox{vec}}
\newcommand{\st}{\mbox{s.t.}}
\newcommand{\grad}{\nabla}
\newcommand{\qsection}[1]{\vspace{4mm} \noindent \textbf{#1:}}
%
%
\newcommand{\etal}{\emph{et al.} }
\newcommand{\ie}{\emph{i.e.} }
\newcommand{\eg}{\emph{e.g.} }
\newcommand{\wrt}{w.r.t. }
\newcommand{\vs}{\emph{vs.} }
%

\newcommand{\red}{\color{red}}
\newcommand{\black}{\color{black}}

\begin{abstract}
The Lucas \& Kanade (LK) algorithm is the method of choice for efficient dense image and object alignment. The approach is efficient as it attempts to model the connection between appearance and geometric displacement through a linear relationship that assumes independence across pixel coordinates. A drawback of the approach, however, is its generative nature. Specifically, its performance is tightly coupled with how well the linear model can synthesize appearance from geometric displacement, even though the alignment task itself is associated with the inverse problem. In this paper, we present a new approach, referred to as the Conditional LK algorithm, which: (i) directly learns linear models that predict geometric displacement as a function of appearance, and (ii) employs a novel strategy for ensuring that the generative pixel independence assumption can still be taken advantage of. We demonstrate that our approach exhibits superior performance to classical generative forms of the LK algorithm. Furthermore, we demonstrate its comparable performance to state-of-the-art methods such as the Supervised Descent Method with substantially less training examples, as well as the unique ability to ``swap'' geometric warp functions without having to retrain from scratch. Finally, from a theoretical perspective, our approach hints at possible redundancies that exist in current state-of-the-art methods for alignment that could be leveraged in vision systems of the future. 

\end{abstract}


\section{Introduction}
The Lucas \& Kanade (LK) algorithm \cite{lucas1981iterative} has been a popular approach for tackling dense alignment problems for images and objects. At the heart of the algorithm is the assumption that an approximate linear relationship exists between pixel appearance and geometric displacement. Such a relationship is seldom exactly linear, so a linearization process is typically repeated until convergence. Pixel intensities are not deterministically differentiable with respect to geometric displacement; instead, the linear relationship must be established stochastically through a learning process. One of the most notable properties of the LK algorithm is how efficiently this linear relationship can be estimated. This efficiency stems from the assumption of independence across pixel coordinates - the parameters describing this linear relationship are classically referred to as image gradients. In practice, these image gradients are estimated through finite differencing operations. Numerous extensions and variations upon the LK algorithm have subsequently been explored in literature~\cite{baker2004lucas}, and recent work has also demonstrated the utility of the LK framework~\cite{antonakos2015feature,bristow14,DBLP:journals/corr/AlismailBL16} using classical dense descriptors such as dense SIFT~\cite{lowe2004distinctive}, HOG~\cite{dalal2005histograms}, and LBP~\cite{ojala2002multiresolution}.

A drawback to the LK algorithm and its variants, however, is its generative nature. Specifically, it attempts to synthesize, through a linear model, how appearance changes as a function of geometric displacement, even though its end goal is the inverse problem. Recently, Xiong \& De la Torre~\cite{xiong2013supervised,DBLP:journals/corr/XiongT14,xiong2015global} proposed a new approach to image alignment known as the Supervised Descent Method (SDM). SDM shares similar properties with the LK algorithm as it also attempts to establish the relationship between appearance and geometric displacement using a sequence of linear models. One marked difference, however, is that SDM directly learns how geometric displacement changes as a function of appearance. This can be viewed as estimating the conditional likelihood function~$p(\y | \x)$, where~$\y$~and~$\x$ are geometric displacement and appearance respectively.  As reported in literature~\cite{jebara2001discriminative} (and also confirmed by our own experiments in this paper), this can lead to substantially improved performance over classical LK as the learning algorithm is focused directly on the end goal (\ie estimating geometric displacement from appearance).

Although it exhibits many favorable properties, SDM also comes with disadvantages. Specifically, due to its non-generative nature, SDM cannot take advantage of the pixel independence assumption enjoyed through classical LK (see Section~\ref{sec:conditionalLK} for a full treatment on this asymmetric property). Instead, it needs to model full dependence across all pixels, which requires: (i) a large amount of training data, and (ii) the requirement of adhoc regularization strategies in order to avoid a poorly conditioned linear system. Furthermore, SDM does not utilize prior knowledge of the type of geometric warp function being employed (\eg similarity, affine, homography, point distribution model, etc.), which further simplifies the learning problem in classical LK.

In this paper, we propose a novel approach which, like SDM, attempts to learn a linear relationship between geometric displacement directly as a function of appearance. However, unlike SDM, we enforce that the pseudo-inverse of this linear relationship enjoys the generative independence assumption across pixels while utilizing prior knowledge of the parametric form of the geometric warp. We refer to our proposed approach as the Conditional LK algorithm. Experiments demonstrate that our approach achieves comparable, and in many cases better, performance to SDM across a myriad of tasks with substantially less training examples. We also show that our approach does not require any adhoc regularization term, and it exhibits a unique property of being able to ``swap'' the type of warp function being modeled (\eg replace a homography with an affine warp function) without the need to retrain. Finally, our approach offers some unique theoretical insights into the redundancies that exist when attempting to learn efficient object/image aligners through a conditional paradigm. 

\textbf{Notations.} We define our notations throughout the paper as follows: lowercase boldface symbols (\eg $\x$) denote vectors, uppercase boldface symbols (\eg $\R$) denote matrices, and uppercase calligraphic symbols (\eg $\I$) denote functions. We treat images as a function of the warp parameters, and we use the notations $\I(\x) : \Real^2 \to \Real^K$ to indicate sampling of the~$K$-channel image representation at subpixel location~$\x = [x,y]^{\top}$. Common examples of multi-channel image representations include descriptors such as dense SIFT, HOG and LBP. We assume~$K = 1$ when dealing with raw grayscale images. 


\section{The Lucas \& Kanade Algorithm}
At its heart, the Lucas \& Kanade (LK) algorithm utilizes the assumption that, 
\begin{equation}
\I(\x + \Delta \x) \approx \I(\x) + \nabla \I(\x) \Delta \x \;\;. 
\label{Eq:approx}
\end{equation}
where~$\I(\x) : \Real^{2} \to \Real^{K}$ is the image function representation and~$\nabla \I(\x): \Real^{2} \to \Real^{K \times 2}$ is 
the image gradient function at pixel coordinate~$\x = [x,y]$. In most instances, a useful image gradient function $\nabla \I(\x)$ can be efficiently estimated through finite differencing operations. 
An alternative strategy is to treat the problem of gradient estimation as a per-pixel linear regression problem, where pixel intensities are samples around a neighborhood in order to ``learn'' the image gradients~\cite{bristow14}. A focus of this paper is to explore this idea further by examining more sophisticated conditional learning objectives for learning image gradients. 

For a given geometric warp function~$\mathcal{W}\{\x;\p\}: \Real^{2} \to \Real^{2}$ parameterized by the warp parameters~$\p \in \Real^{P}$, one can thus express the classic LK algorithm as minimizing the sum of squared differences (SSD) objective, 
\begin{equation}
\min_{\deltap} \sum_{d=1}^{D} \eucsqnorm{\I(\mathcal{W}\{\x_{d}; \p\}) + \nabla \I(\mathcal{W}\{\x_{d}; \p\}) \frac{\partial \mathcal{W}(\x_{d}; \p)}{\partial \p^{\top}} \Delta \p - \T(\x_{d})} ,
\label{Eq:LK}
\end{equation} 
which can be viewed as a quasi-Newton update. The parameter~$\p$ is the initial warp estimate, ~$\Delta \p$ is the warp update being estimated, and~$\T$ is the template image we desire to align the source image~$\I$ against. The pixel coordinates~$\{ \x_{d} \}_{d=1}^{D}$ are taken with respect to the template image's coordinate frame, and~$\frac{\partial \mathcal{W}(\x; \p)}{\partial \p^{\top}}: \Real^{2} \to \Real^{2 \times P}$ is the warp Jacobian. After solving Equation~\ref{Eq:LK}, the current warp estimate has the following additive update, 
\begin{equation}
\p \leftarrow \p + \Delta \p\;. 
\label{Eq:update}
\end{equation}
As the relationship between appearance and geometric deformation is not solely linear, Equations~\ref{Eq:LK} and~\ref{Eq:update} must be applied iteratively until convergence is achieved.

\subsubsection{Inverse compositional fitting.}
The canonical LK formulation presented in the previous section is sometimes referred to as the forwards additive (FA) algorithm~\cite{baker2004lucas}. 
A fundamental problem with the forwards additive approach is that it requires recomputing the image gradient and warp Jacobian in each iteration, greatly impacting 
computational efficiency. Baker and Matthews~\cite{baker2004lucas} devised a computationally efficient extension to forwards additive LK, which they refer to as the inverse compositional (IC) algorithm. The IC-LK algorithm attempts to iteratively solve the objective
\begin{equation}
\min_{\deltap} \sum_{d=1}^{D} \eucsqnorm{\I(\mathcal{W}\{\x_{d}; \p\}) - \T(\x_{d}) - \nabla \T(\x_{d}) \frac{\partial \mathcal{W}(\x_{d}; \0)}{\partial \p^{\top} } \Delta \p} ,
\label{Eq:IC}
\end{equation} 
followed by the inverse compositional update
\begin{equation}
\p \leftarrow \p \circ (\Delta \p)^{-1} ,
\end{equation} 
where we have abbreviated the notation $\circ$ to be the composition of warp functions parametrized by $\p$, and $(\Delta \p)^{-1}$ to be the parameters of the inverse warp function parametrized by $\deltap$.
We can express Equation~\ref{Eq:IC} in vector form as
\begin{equation}
\min_{\deltap} \eucsqnorm{\I(\p) - \T(\0) - \W \Delta \p} ,
\label{Eq:VIC}
\end{equation} 
where,
\begin{equation*}
\W = \begin{bmatrix} \nabla \T(\x_{1}) & \hdots & 0 \\ \vdots & \ddots & \vdots \\ 0 & \hdots & \nabla \T(\x_{D}) \end{bmatrix} 
\begin{bmatrix}
\frac{\partial \mathcal{W}(\x_{1}; \0)}{\partial \p^{\top} } \\
\vdots \\
\frac{\partial \mathcal{W}(\x_{D}; \0)}{\partial \p^{\top} }
\end{bmatrix}
\end{equation*}
and
\begin{equation*}
\I(\p) = \begin{bmatrix} \I(\mathcal{W}\{\x_{1};\p\}) \\ \vdots \\ \I(\mathcal{W}\{\x_{D};\p\}) \end{bmatrix}, 
\quad \T(\0) = \begin{bmatrix} \T(\mathcal{W}\{\x_{1};\0\}) \\ \vdots \\ \T(\mathcal{W}\{\x_{D}; \0 \}) \end{bmatrix} .
\end{equation*}
Here, $\p = \0$ is considered the identity warp (\ie $\mathcal{W}\{\x;\0\} = \x$). It is easy to show that the solution to Equation~\ref{Eq:VIC} is given by 
\begin{equation}
\Delta \p = \R[\I(\p) - \T(\0)] ,
\end{equation}
where~$\R = \W^{\dagger}$.  The superscript $\dagger$ denotes the Moore-Penrose pseudo-inverse operator. The IC form of the LK algorithm comes with a great advantage: the gradients~$\nabla \T(\x)$ and warp Jacobian~$\frac{\partial \mathcal{W}(\x; \0)}{\partial \p^{\top} }$ are evaluated at the identity warp $\p = \0$, regardless of the iterations and the current state of $\p$. This means that~$\R$ remains constant across all iterations, making it advantageous over other variants in terms of computational complexity. For the rest of this paper, we shall focus on the IC form of the LK algorithm. 

\section{Supervised Descent Method} \label{sec:SDM}
Despite exhibiting good performance on many image alignment tasks, the LK algorithm can be problematic to use when there is no specific template image~$\T$ to align against. For many applications, one may be given just an ensemble of $M$ ground-truth images and warps~$\{ \I_{m}, \p_{m} \}_{m=1}^{M}$ of the object of interest. If one has prior knowledge of the distribution of warp displacements to be encountered, one can synthetically generate $N$ examples to form a much larger set~$\S = \{ \Delta \p_{n}, \I_{n}(\p_{n} \circ \Delta \p_{n}) \}_{n=1}^{N}$ to learn from, where~$N \gg M$. In these circumstances, a strategy recently put forward known as the Supervised Descent Method (SDM)~\cite{xiong2013supervised} has exhibited state-of-the-art performance across a number of alignment tasks, most notably facial landmark alignment. The approach attempts to directly learn a regression matrix that minimizes the following SSD objective, 
\begin{equation}
\min_{\R} \sum_{n \in \S} \eucsqnorm{\Delta \p_{n} - \R [\I_{n} (\p_{n} \circ \Delta \p_{n} ) - \T(\0)]} + \Omega(\R) \;\;. 
\label{Eq:SDM}
\end{equation}
The template image $\TO$ can be learned either with $\R$ directly or by taking it to be $\frac{1}{N} \sum_{n \in \S} \I(\p_{n})$, the average of ground-truth images~\cite{xiong2015global}.

\subsubsection{Regularization.}
$\Omega$ is a regularization function used to ensure that the solution to~$\R$ is unique. To understand the need for this regularization, one can reform Equation~\ref{Eq:SDM} in matrix form as
\begin{equation}
\min_{\R} \frobnorm{\Y - \R \X} + \Omega(\R) ,
\label{Eq:MSDM}
\end{equation} 
where
\begin{align*}
\Y &= \begin{bmatrix} \Delta \p_{1}, \ldots, \Delta \p_{N} \end{bmatrix}, \mbox{and} \\
\X &= \begin{bmatrix} \I(\p_{1} \circ \Delta \p_{1}) - \T(\0), \ldots, \I(\p_{N} \circ \Delta \p_{N} ) - \T(\0) \end{bmatrix}\;\; . 
\end{align*}
Here, $\left\|\cdot\right\|_F$ indicates the matrix Frobenius norm. Without the regularization term $\Omega(\R)$, the solution to Equation~\ref{Eq:MSDM} is ~$\R = \Y \X^{\top} (\X \X^{\top})^{-1}$. It is understood in literature that raw pixel representations of natural images stem from certain frequency spectrums~\cite{simoncelli2001natural} that leads to an auto-covariance matrix $\X \X^{\top}$ which is poorly conditioned in nearly all circumstances. It has been demonstrated~\cite{simoncelli2001natural} that this property stems from the fact that image intensities in natural images are highly correlated in close spatial proximity, but this dependence drops off as a function of spatial distance.

In our experiments, we have found that~$\X \X^{\top}$ is always poorly conditioned even when utilizing other image representations such as dense SIFT, HOG, and LBP descriptors. As such, it is clear that some sort of regularization term is crucial for effective SDM performance. As commonly advocated and practiced, we employed a weighted Tikhonov penalty term~$\Omega(\R) = \lambda || \R ||_{F}^{2}$, where~$\lambda$ controls the weight of the regularizer. We found this choice to work well in our experiments. 

\subsubsection{Iteration-specific Regressors.}
Unlike the IC-LK approach, which employs a single regressor/template pair~$\{\R, \TO\}$ to be applied iteratively until convergence, SDM learns a set of regressor/template pairs $\{ \R^{(l)}, \mathcal{T}^{(l)}(\0) \}_{l=1}^{L}$ for each iteration~$l = 1 : L$ (sometimes referred to as layers). On the other hand, like the IC-LK algorithm, these regressors are precomputed in advance and thus are independent of the current image and warp estimate. As a result, SDM is computationally efficient just like IC-LK. The regressor/template pair~$\{ \R^{(l)}, \mathcal{T}^{(l)}(\0) \}$ is learned from the synthetically generated set~$\S^{(l)}$ within Equation~\ref{Eq:SDM}, which we define to be 
\begin{equation}
\S^{(l)} = \{ \Delta \p^{(l)}_{n}, \I(\p_{n} \circ \Delta \p^{(l)}_{n}) \}_{n=1}^{N} ,
\label{Eq:Set}
\end{equation}
where
\begin{equation}
\Delta \p^{(l+1)} \leftarrow \R^{(l)} \left[\I\left(\p \circ (\Delta \p^{(l)})^{-1} \right) - \T(\0) \right] \;. 
\label{Eq:SDMapp}
\end{equation}
For the first iteration~($l=1$), the warp perturbations are generated from a pre-determined random distribution; for every subsequent iteration, the warp perturbations are re-sampled from the same distribution to ensure each iteration's regressor does not overfit. Once learned, SDM is applied by employing Equation~\ref{Eq:SDMapp} in practice. 

\subsubsection{Inverse Compositional Warps.}
It should be noted that there is nothing in the original treatment~\cite{xiong2013supervised} on SDM that limits it to compositional warps. In fact, the original work employing facial landmark alignment advocated an additive update strategy. In this paper, however, we have chosen to employ inverse compositional warp updates as: (i) we obtained better results for our experiments with planar warp functions, (ii) we observed almost no difference in performance for non-planar warp functions such as those involved in face alignment, and (iii) it is only through the employment of inverse compositional warps within the LK framework that a firm theoretical motivation for fixed regressors can be entertained.  Furthermore, we have found that keeping a close mathematical relationship to the IC-LK algorithm is essential for the motivation of our proposed approach. 

\section{The Conditional Lucas \& Kanade Algorithm} \label{sec:conditionalLK}
Although enjoying impressive results across a myriad of image alignment tasks, SDM does have disadvantages when compared to IC-LK. First, it requires large amounts of synthetically warped image data. Second, it requires the utilization of an adhoc regularization strategy to ensure good condition of the linear system. Third, the mathematical properties of the warp function parameters being predicted is ignored. Finally, it reveals little about the actual degrees of freedom necessary in the set of regressor matrices being learned through the SDM process. 

In this paper, we put forward an alternative strategy for directly learning a set of iteration-specific regressors, 
\begin{eqnarray} \label{Eq:CLK}
\min_{\nabla \T(\0)} & \sum_{n \in \S} & \eucsqnorm{\Delta \p_{n} - \R[\I(\p_{n} \circ \Delta \p_{n} ) - \T(\0)]}  \\
 & \mbox {s.t. } & \R = \left( \begin{bmatrix} \nabla \T(\x_{1}) & \hdots & 0 \\ \vdots & \ddots & \vdots \\ 0 & \hdots & \nabla \T(\x_{D}) \end{bmatrix} 
\begin{bmatrix}
\frac{\partial \mathcal{W}(\x_{1}; \0)}{\partial \p^{\top} } \\
\vdots \\
\frac{\partial \mathcal{W}(\x_{D}; \0)}{\partial \p^{\top} } \nonumber
\end{bmatrix} \right)^{\dagger} ,
\end{eqnarray}
where
\begin{equation}
\nabla \T(\0) = \begin{bmatrix} \nabla \T (\x_{1}) \\ \vdots \\ \nabla \T(\x_{D}) \end{bmatrix} \;\;. \nonumber
\end{equation}
At first glance, this objective may seem strange, as we are proposing to learn template ``image gradients'' ~$\nabla \T(\0)$ within a conditional objective. As previously discussed in~\cite{bristow14}, this idea deviates from the traditional view of what image gradients are - parameters that are derived from heuristic finite differencing operations. In this paper, we prefer to subscribe to the alternate view that image gradients are simply weights that can be, and should be, learned from data. The central motivation for this objective is to enforce the parametric form of the generative IC-LK form through a conditional objective.

An advantage of the Conditional LK approach is the reduced number of model parameters. Comparing the model parameters of Conditional LK ($\nabla \T(\0) \in \Real^{KD \times 2}$) against SDM ($\R \in \Real^{P \times KD}$), there is a reduction in the degrees of freedom needing to be learned for most warp functions where~$P > 2$ . More fundamentally, however, is the employment of the generative pixel independence assumption described originally in Equation~\ref{Eq:approx}. This independence assumption is useful as it ensures that a unique~$\R$ can be found in Equation~\ref{Eq:CLK} without any extra penalty terms such as Tikhonov regularization. In fact, we propose that the sparse matrix structure of image gradients within the psuedo-inverse of~$\R$ acts as a much more principled form of regularization than those commonly employed within the SDM framework. 

A further advantage of our approach is that, like the IC-LK framework, it utilizes prior knowledge of the warp Jacobian function~$\frac{\partial \mathcal{W}(\x; \0)}{\partial \p^{\top}}$ during the estimation of the regression matrix~$\R$. Our insight here is that the estimation of the regression matrix~$\R$ using a conditional learning objective should be simplified (in terms of the degrees of freedom to learn) if one had prior knowledge of the deterministic form of the geometric warp function. 

A drawback to the approach, in comparison to both the SDM and IC-LK frameworks, is the non-linear form of the objective in Equation~\ref{Eq:CLK}. This requires us to resort to non-linear optimization methods, which are not as straightforward as linear regression solutions. However, as we discuss in more detail in the experimental portion of this paper, we demonstrate that a Levenberg-Marquardt optimization strategy obtains good results in nearly all circumstances. Furthermore, compared to SDM, we demonstrate good solutions can be obtained with significantly smaller numbers of training samples. 

\subsubsection{Iteration-specific Regressors.}
As with SDM, we assume we have an ensemble of images and ground-truth warps~$\{ \I_{m}, \p_{m} \}_{m=1}^{M}$ from which a much larger set of synthetic examples can be generated~$\S = \{ \Delta \p_{n}, \I_{n}(\p_{n} \circ \Delta \p_{n}) \}_{n=1}^{N}$, where~$N \gg M$. Like SDM, we attempt to learn a set of regressor/template pairs $\{ \R^{(l)}, \mathcal{T}^{(l)}(\0) \}_{l=1}^{L}$ for each iteration~$l = 1 : L$. The set~$\S^{(l)}$ of training samples is derived from Equations~\ref{Eq:Set} and~\ref{Eq:SDMapp} for each iteration. Once learned, the application of these iteration-specific regressors is identical to SDM. 

\subsubsection{Pixel Independence Asymmetry.}
A major advantage of the IC-LK framework is that it assumes generative independence across pixel coordinates (see Equation~\ref{Eq:approx}). A natural question to ask is: could not one predict geometric displacement (instead of appearance) directly across independent pixel coordinates?

The major drawback to employing such strategy is its ignorance of the well-known ``aperture problem''~\cite{marr1982vision} in computer vision (\eg the motion of an image patch containing a sole edge cannot be uniquely determined due to the ambiguity of motion along the edge). As such, it is impossible to ask any predictor (linear or otherwise) to determine the geometric displacement of all pixels within an image while entertaining an independence assumption.
The essence of our proposed approach is that it circumvents this issue by enforcing global knowledge of the template's appearance across all pixel coordinates, while entertaining the generative pixel independence assumption that has served the LK algorithm so well over the last three decades. 

\begin{figure}[t]
\includegraphics[width=\textwidth]{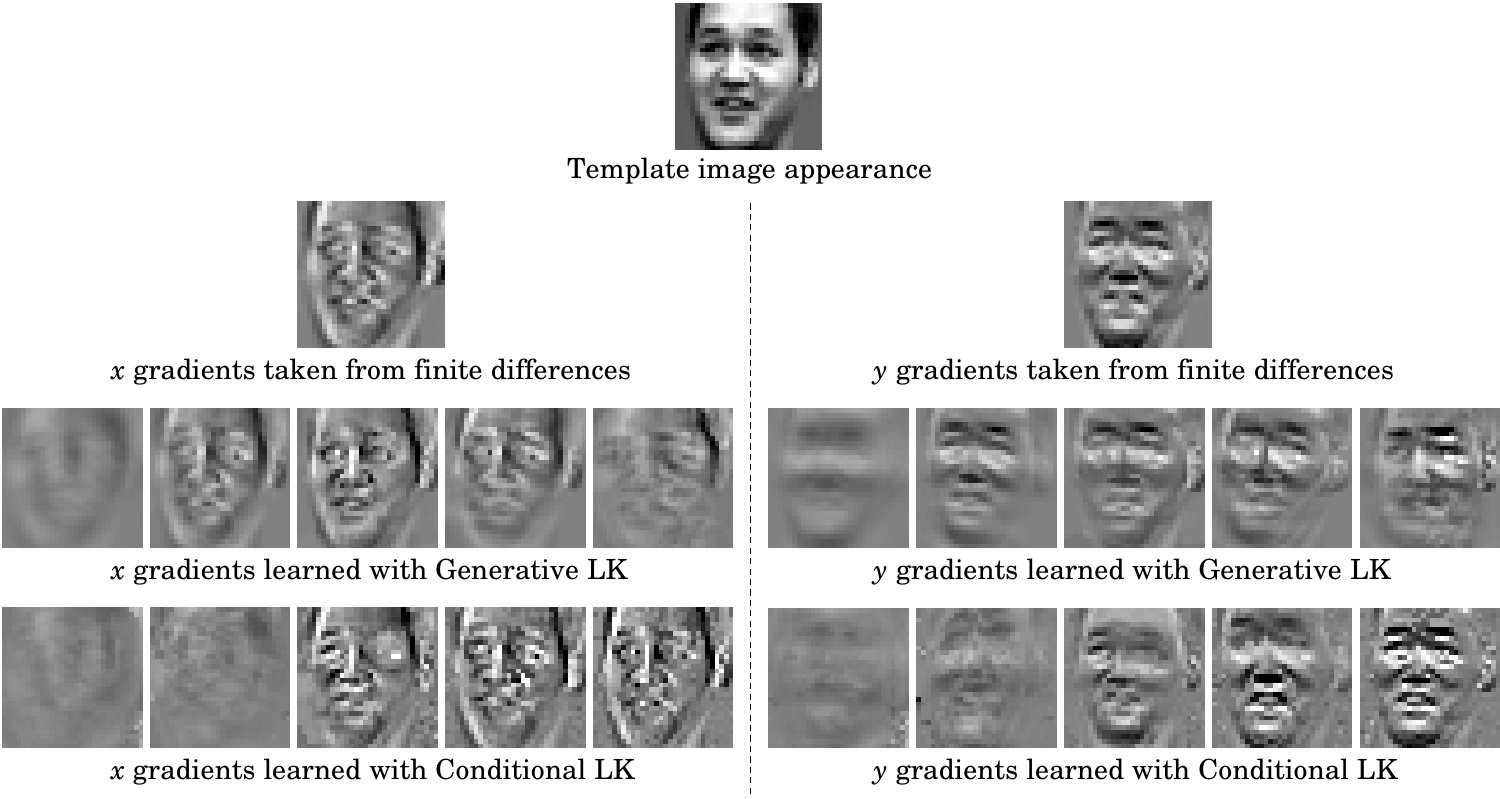} 
\centering
\caption{Visualization of the learned image gradients for LK from layers 1 (left) to 5 (right).}
\label{fig:visualizegradients}
\end{figure}

\subsubsection{Generative LK.}
For completeness, we will also entertain a generative form of our 
objective in Equation~\ref{Eq:CLK}, where we instead learn ``image gradients'' that predict generative appearance as a function of geometric displacement, formulated as
\begin{eqnarray}
\label{Eq:GLK}
\min_{\nabla \T(\0)} & \sum_{n \in \S} & \eucsqnorm{ \I(\p_{n} \circ \Delta \p_{n} ) - \T(\0) - \W \Delta \p_{n} }  \\
 & \mbox {s.t. } & \W = \begin{bmatrix} \nabla \T(\x_{1}) & \hdots & 0 \\ \vdots & \ddots & \vdots \\ 0 & \hdots & \nabla \T(\x_{D}) \end{bmatrix} 
\begin{bmatrix}
\frac{\partial \mathcal{W}(\x_{1}; \0)}{\partial \p^{\top} } \\
\vdots \\
\frac{\partial \mathcal{W}(\x_{D}; \0)}{\partial \p^{\top} } \nonumber
\end{bmatrix}.
\end{eqnarray}
Unlike our proposed Conditional LK, the objective in Equation~\ref{Eq:GLK} is linear and directly solvable. Furthermore, due to the generative pixel independence assumption, the problem can be broken down into~$D$ independent sub-problems. The Generative LK approach is trained in an identical way to SDM and Conditional LK, where iteration-specific regressors are learned from a set of synthetic examples~$\S = \{ \Delta \p_{n}, \I_{n}(\p_{n} \circ \Delta \p_{n}) \}_{n=1}^{N}$.

Figure \ref{fig:visualizegradients} provides an example of visualizing the gradients learned from the Conditional LK and Generative LK approaches. It is worthwhile to note that the Conditional LK gradients get sharper over regression iterations, while it is not necessarily the case for Generative LK. The rationale for including the Generative LK form is to highlight the importance of a conditional learning approach, and to therefore justify the added non-linear complexity of the objective in Equation~\ref{Eq:CLK}. 


\section{Experiments}

In this section, we present results for our approach across three diverse tasks: (i) planar image alignment, (ii) planar template tracking, and (iii) facial model fitting. We also investigate the utility of our approach across different image representations such as raw pixel intensities and dense LBP descriptors.  

\subsection{Planar Image Alignment}
\subsubsection{Experimental settings.}

\begin{figure}[t]
\includegraphics[width=1\textwidth]{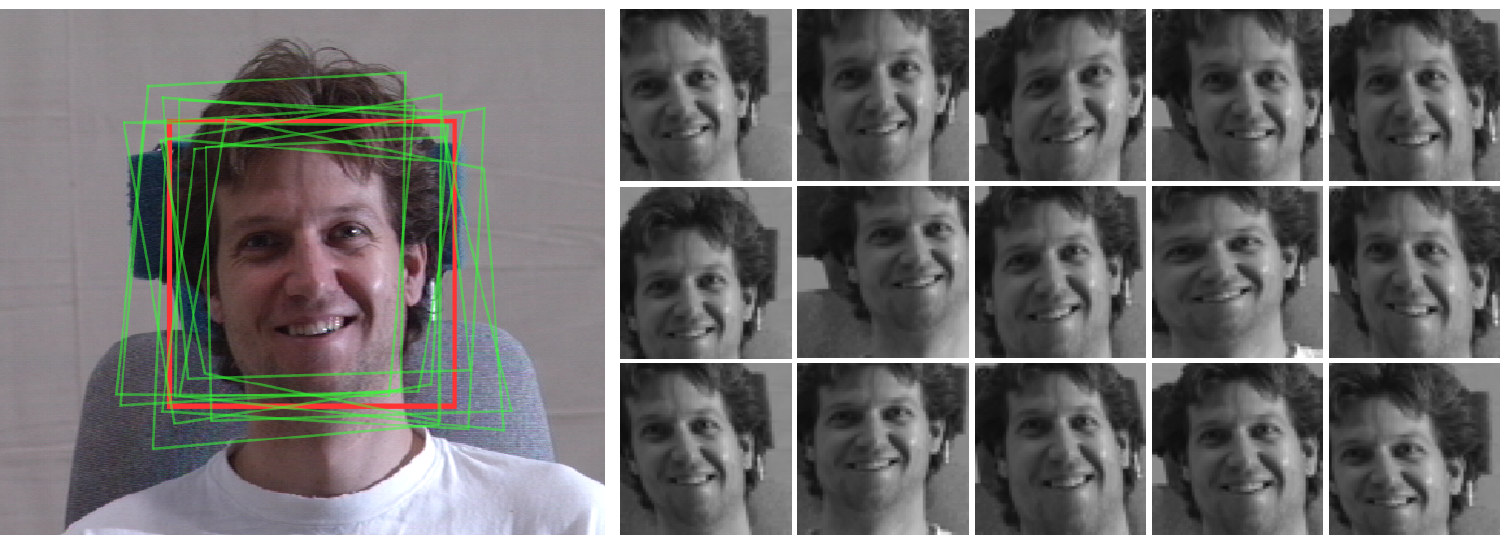} 
\centering
\caption{Visualization of the perturbed samples $\S = \{ \Delta \p_{n}, \I_{n}(\p_{n} \circ \Delta \p_{n}) \}_{n=1}^{N}$ used for training the SDM, Conditional LK, and Generative LK methods. Left: the original source image, where the red box is the ground truth and the green boxes are perturbed for training. Right: examples of the synthesized training samples.}
\label{fig:visualizetraining}
\end{figure}

In this portion of our experiments, we will be utilizing a subsection of the Multi-PIE~\cite{gross2010multi} dataset. For each image, we denote a~$20 \times 20$ image~$\I(\p)$ with ground-truth warp~$\p$ rotated, scaled and translated around hand-labeled locations. For the IC-LK approach, this image is then employed as the template~$\T(\0)$. For the SDM, Conditional LK and Generative LK methods, a synthetic set of geometrically perturbed samples~$\S$ are generated $\S = \{ \Delta \p_{n}, \I_{n}(\p_{n} \circ \Delta \p_{n}) \}_{n=1}^{N}$.

We generate the perturbed samples by adding i.i.d. Gaussian noise of standard deviation $\sigma$ to the four corners of the ground-truth bounding box as well as an additional translational noise from the same distribution, and then finally fitting the perturbed box to the warp parameters $\deltap$. In our experiments, we choose $\sigma = 1.2$ pixels. Figure \ref{fig:visualizetraining} shows an example visualization of the training procedure as well as the generated samples. For SDM, a Tikhonov regularization term is added to the training objective as described in Section \ref{sec:SDM}, and the penalty factor $\lambda$ is chosen by evaluating on a separate validation set; for Conditional LK, we use Levenberg-Marquardt to optimize the non-linear objective where the parameters are initialized through the Generative LK solution.

\subsubsection{Frequency of Convergence.}
\begin{figure}[t]
\includegraphics[width=0.32\textwidth]{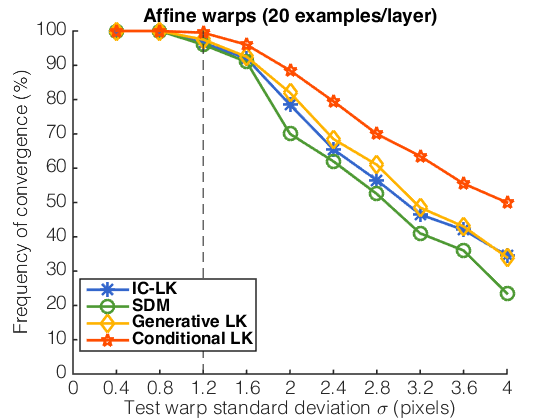} 
\includegraphics[width=0.32\textwidth]{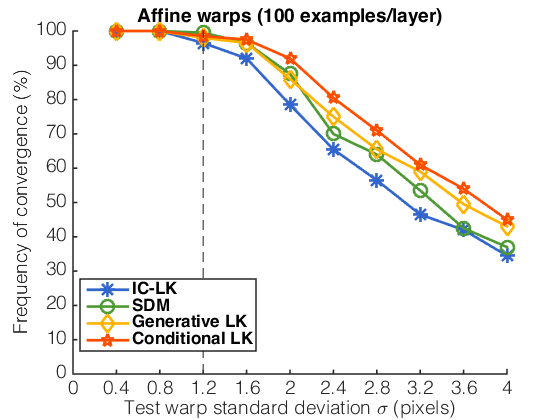} 
\includegraphics[width=0.32\textwidth]{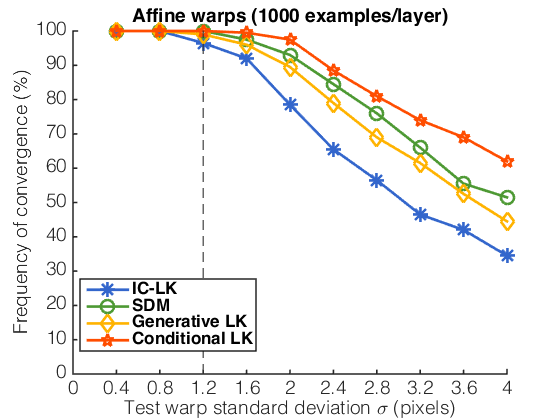} 
\includegraphics[width=0.32\textwidth]{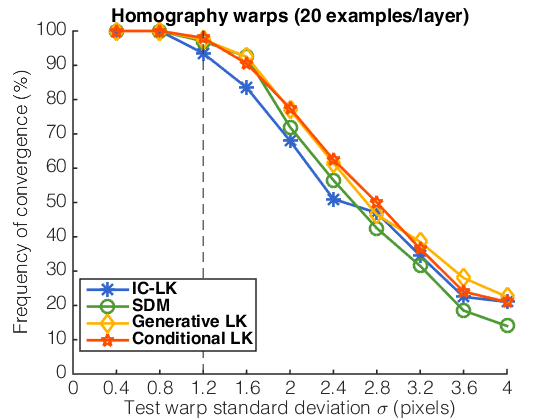} 
\includegraphics[width=0.32\textwidth]{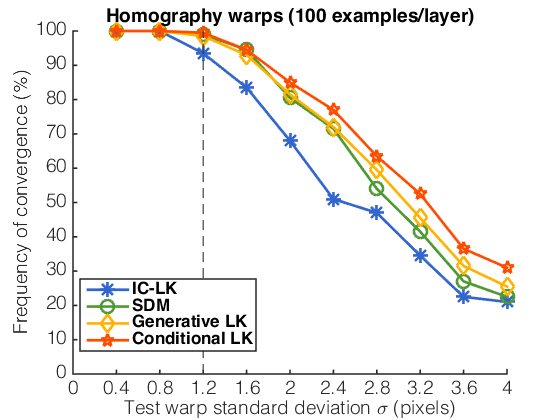} 
\includegraphics[width=0.32\textwidth]{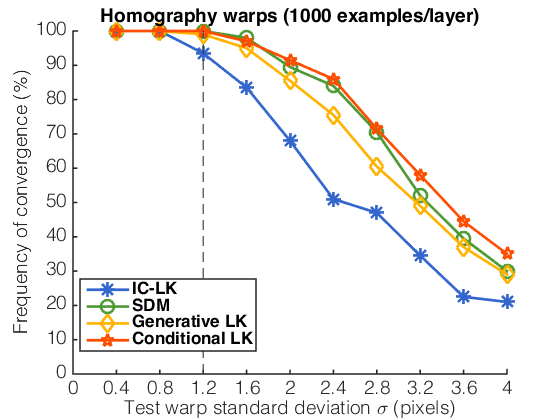} 
\centering
\caption{Frequency of convergence comparison between IC-LK, SDM, Generative LK, and Conditional LK. The vertical dotted line indicates $\sigma$ that they were trained with.}
\label{fig:freqconvg_pixel}
\end{figure}
We compare the alignment performance of the four types of aligners in our discussion: (i) IC-LK, (ii) SDM, (iii) Generative LK, and (iv) Conditional LK. We state 
that convergence is reached when the point RMSE of the four corners of the bounding box is less than one pixel.

Figure \ref{fig:freqconvg_pixel} shows the frequency of convergence tested with both a 2D affine and homography warp function. Irrespective of the planar warping function, our results indicate that Conditional LK has superior convergence properties over the others. This result holds even when the approach is initialized with a warp perturbation that is larger than the distribution it was trained under. The alignment performance of Conditional LK is consistently better in all circumstances, although the advantage of the approach is most noticeable when training with just a few training samples.

\begin{figure}[t]
\includegraphics[width=0.24\textwidth]{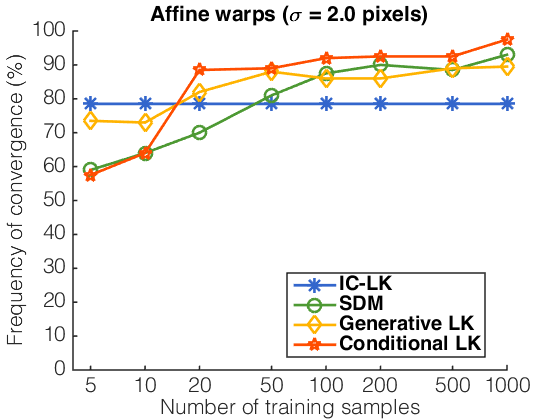} 
\includegraphics[width=0.24\textwidth]{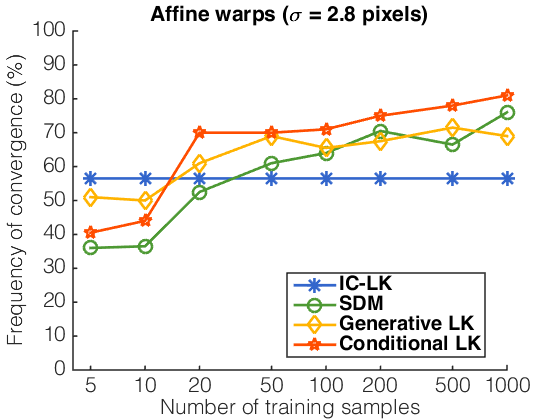} 
\includegraphics[width=0.24\textwidth]{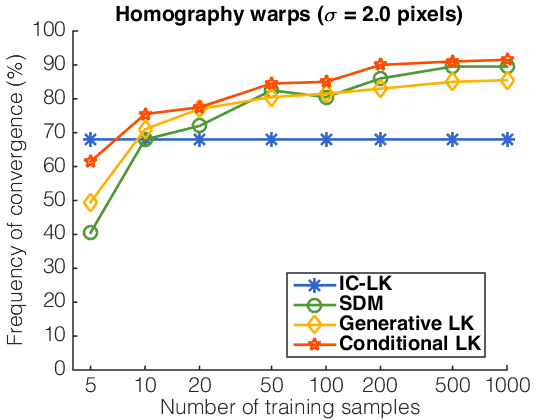} 
\includegraphics[width=0.24\textwidth]{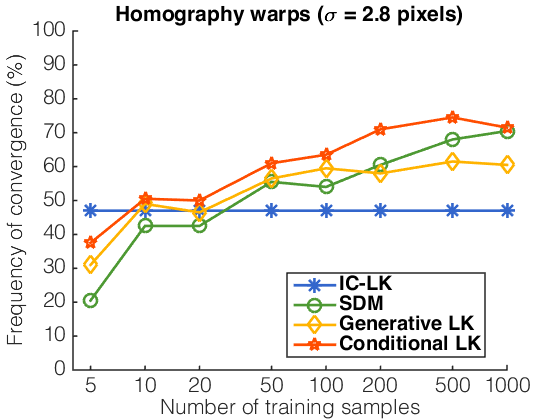} 
\centering
\caption{Frequency of convergence comparison between SDM, Generative LK, and Conditional LK in terms of number of samples trained with.}
\label{fig:freqconvg2_pixel}
\end{figure}
Figure~\ref{fig:freqconvg2_pixel} provides another comparison with respect to the amount of training data learned from. It can be observed that SDM is highly dependent on the amount of training data available, but it is still not able to generalize as well as Conditional LK. This is also empirical proof that incorporating principled priors in Conditional LK is more desirable than adhoc regularizations in SDM.

\subsubsection{Convergence Rate.}

\begin{figure}[t]
\includegraphics[width=0.32\textwidth]{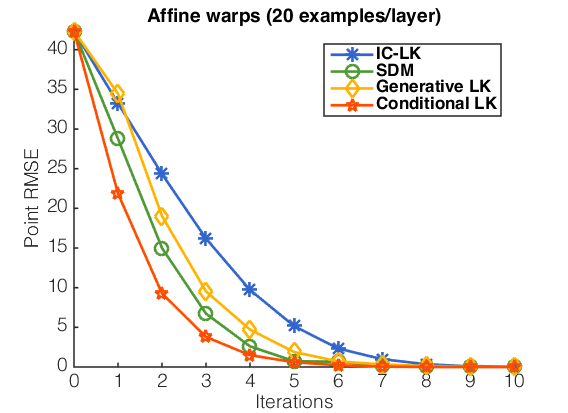} 
\includegraphics[width=0.32\textwidth]{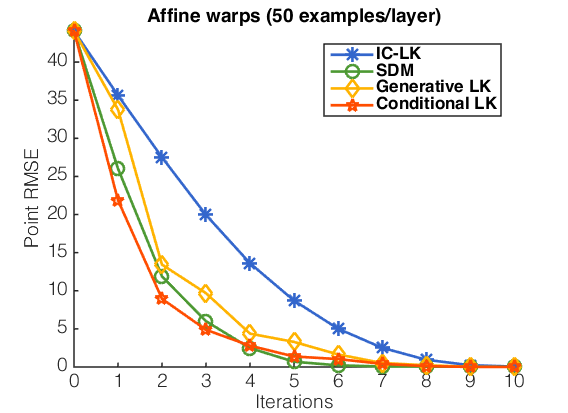} 
\includegraphics[width=0.32\textwidth]{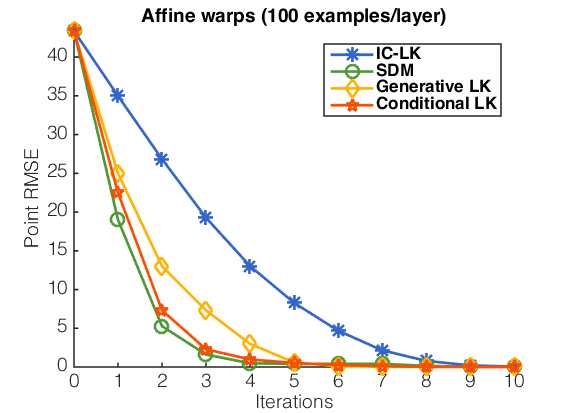} 
\includegraphics[width=0.32\textwidth]{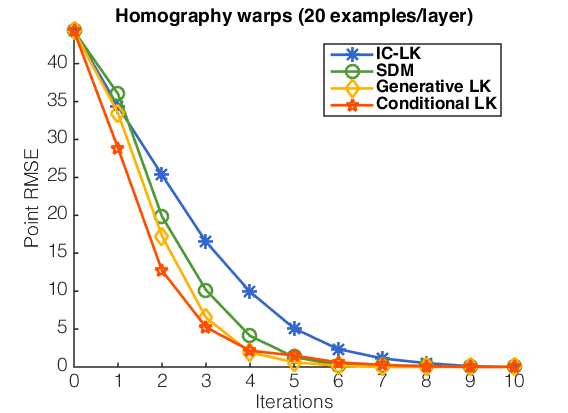} 
\includegraphics[width=0.32\textwidth]{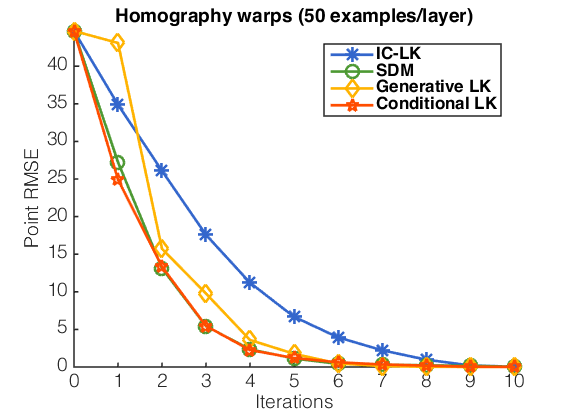} 
\includegraphics[width=0.32\textwidth]{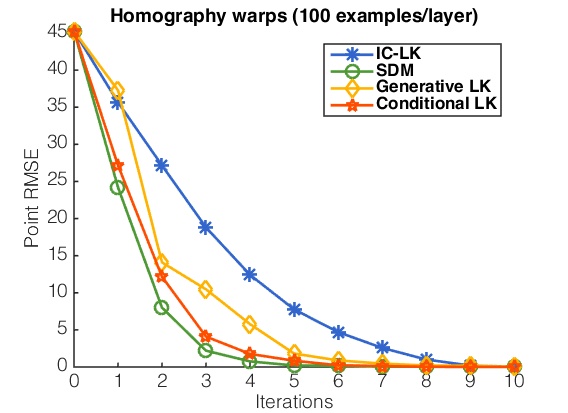} 
\centering
\caption{Convergence rate comparison between IC-LK, SDM, Generative LK, and Conditional LK, averaged from the tests ($\sigma=2.8$) where all four converged in the end.}
\label{fig:convgrate_pixel}
\end{figure}

We also provide some analysis on the convergence speed. To make a fair comparison, we take the average of only those test runs where all regressors converged. Figure \ref{fig:convgrate_pixel} illustrates the convergence rates of different regressors learned from different amounts of training data. The improvement of Conditional LK in convergence speed is clear, especially when little training data is provided. SDM starts to exhibit faster convergence rate when learned from over $100$ examples per layer; however, Conditional LK still surpasses SDM in term of the frequency of final convergence.

\subsubsection{Swapping Warp Functions.}
\begin{figure}[t]
\includegraphics[width=0.32\textwidth]{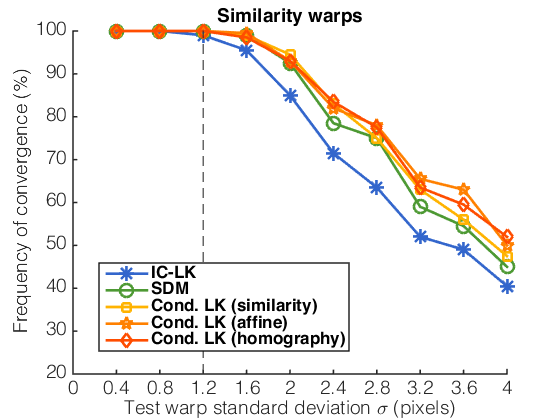} 
\includegraphics[width=0.32\textwidth]{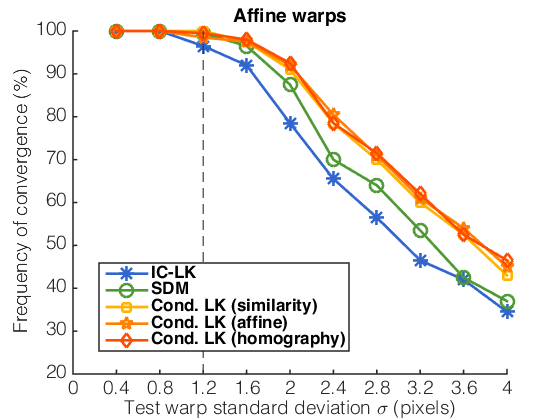} 
\includegraphics[width=0.32\textwidth]{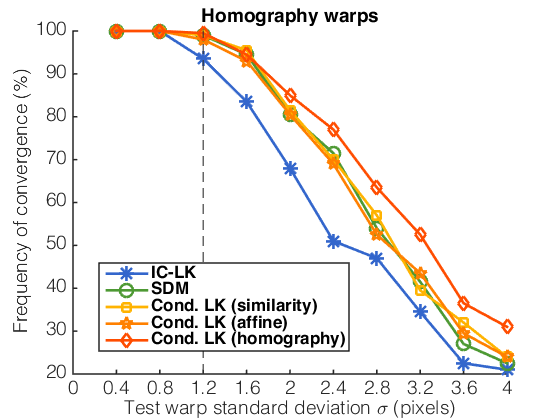} 
\centering
\caption{Frequency of convergence comparison between IC-LK, SDM, and Conditional LK trained with $100$ examples per layer and tested with swapped warp functions. The parentheses indicate the type of warp function trained with.}
\label{fig:swapwarp}
\end{figure}

A unique property of Conditional LK in relation to SDM is its ability to interchange between warp functions after training. Since we are learning image gradients~$\nabla \T(\0)$ for the Conditional LK algorithm, one can essentially choose which warp Jacobian to be employed before forming the regressor~$\R$. Figure \ref{fig:swapwarp} illustrates the effect of Conditional LK learning the gradient with one type of warp function and swapping it with another during testing. We see that whichever warp function Conditional LK is learned with, the learned conditional gradients are also effective on the other and still outperforms IC-LK and SDM.

It is interesting to note that when we learn the Conditional LK gradients using either 2D planar similarity warps ($P=4$) or homography warps ($P=8$), the performance on 2D planar affine warps ($P=6$) is as effective. This outcome leads to an important insight: it is possible to learn the conditional gradients with a simple warp function and replace it with a more complex one afterwards; this can be especially useful when certain types of warp functions (\eg 3D warp functions) are harder to come by.

\subsection{Planar Tracking with LBP Features}
In this section, we show how Conditional LK can be effectively employed with dense multi-channel LBP descriptors where~$K = 8$. First we analyze the 
convergence properties of Conditional LK on the dense LBP descriptors, as we did similarly in the previous section, and then we present an application to robust planar tracking. A full description of the multi-channel LBP descriptors we used in our approach can be found in~\cite{DBLP:journals/corr/AlismailBL16}. 

\begin{figure}[t]
\includegraphics[width=0.32\textwidth]{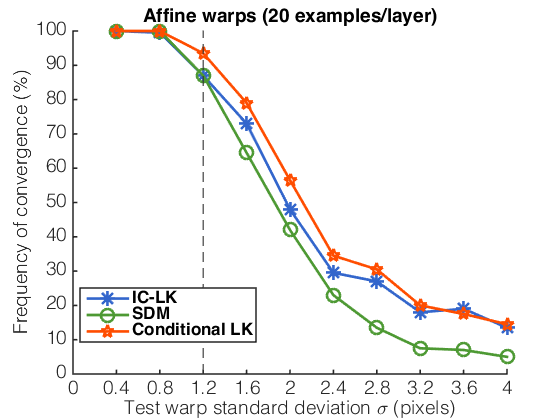} 
\includegraphics[width=0.32\textwidth]{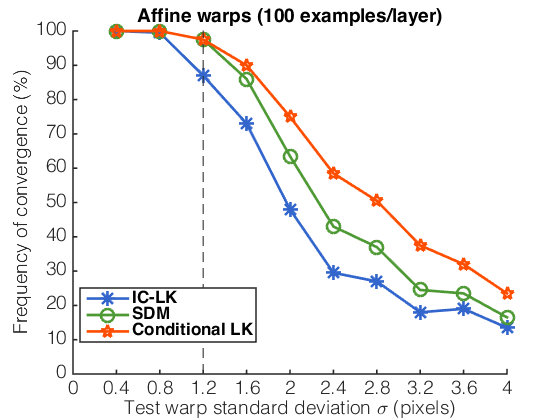} 
\includegraphics[width=0.32\textwidth]{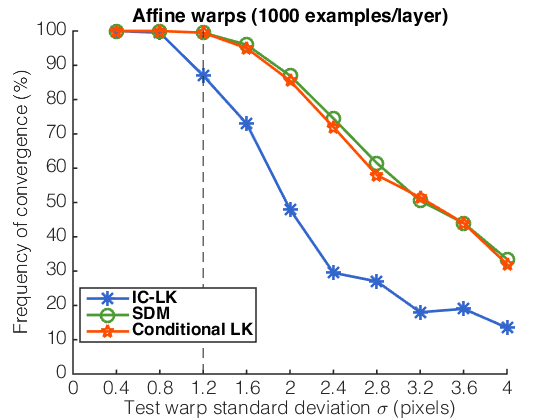} 
\includegraphics[width=0.32\textwidth]{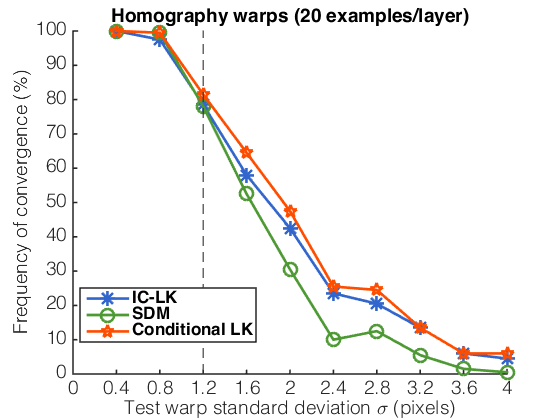} 
\includegraphics[width=0.32\textwidth]{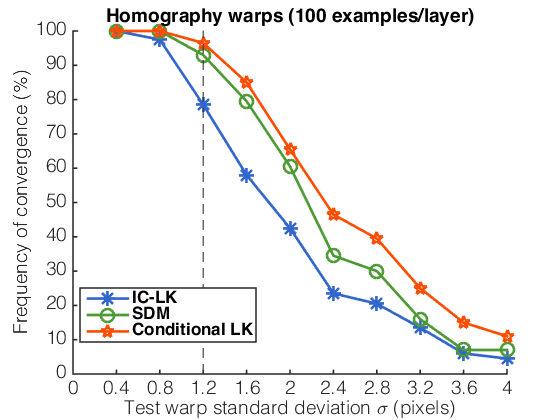} 
\includegraphics[width=0.32\textwidth]{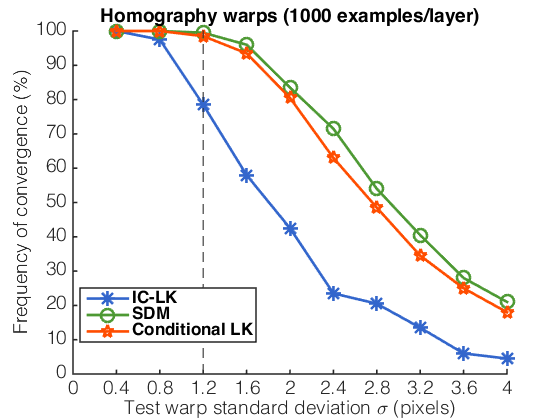} 
\centering
\caption{Frequency of convergence comparison between IC-LK, SDM and Conditional LK with dense binary descriptors. The vertical dotted line indicates $\sigma$ that they were trained with.}
\label{fig:freqconvg_LBP}
\end{figure}

Figure \ref{fig:freqconvg_LBP} provides a comparison of robustness by evaluating the frequency of convergence with respect to the scale of test warps $\sigma$. This suggests that Conditional LK is as effective in the LK framework with multi-channel descriptors: in addition to increasing alignment robustness (which is already a well-understood property of descriptor image alignment), Conditional LK is able to improve upon the sensitivity to initialization with larger warps.

\begin{figure}[t]
\includegraphics[width=0.24\textwidth]{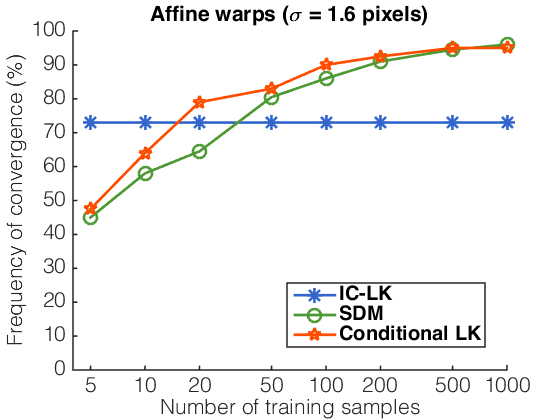} 
\includegraphics[width=0.24\textwidth]{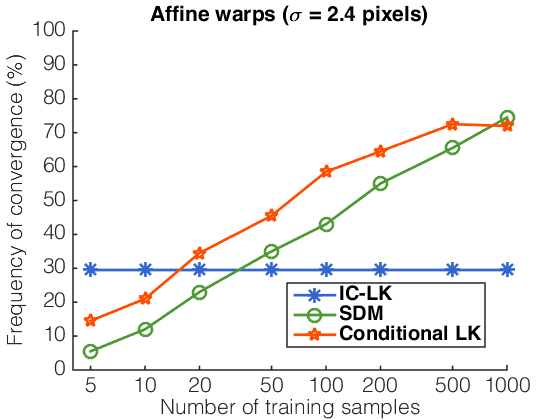} 
\includegraphics[width=0.24\textwidth]{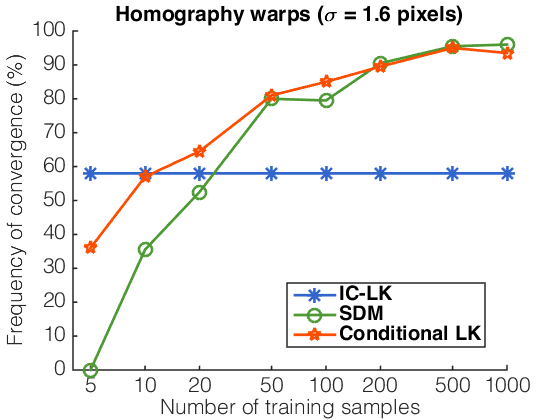} 
\includegraphics[width=0.24\textwidth]{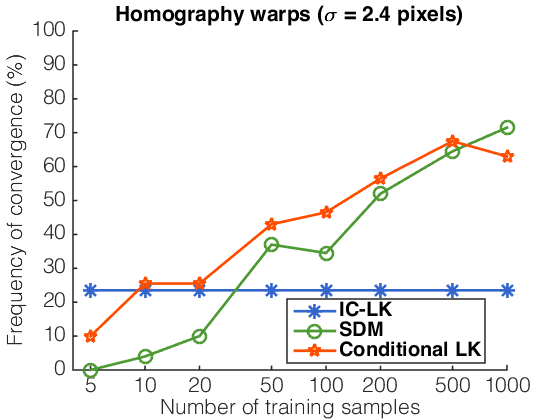} 
\centering
\caption{Frequency of convergence comparison between SDM and Conditional LK with dense binary descriptors in terms of number of samples trained with.}
\label{fig:freqconvg2_LBP}
\end{figure}
Figure \ref{fig:freqconvg2_LBP} illustrates alignment performance as a function of the number of samples used in training. We can see the Conditional LK only requires as few as 20 examples per layer to train a better multi-channel aligner than IC-LK, whereas SDM needs more than 50 examples per iteration-specific regressor. This result again speaks to the efficiency of learning with Conditional LK.

\subsubsection{Low Frame-rate Template Tracking.}
In this experiment, we evaluate the advantage of our proposed approach for the task of low frame-rate template tracking. Specifically, we borrow a similar experimental setup to Bit-Planes~\cite{DBLP:journals/corr/AlismailBL16}. LBP-style dense descriptors are ideal for this type of task as their computation is computationally feasible in real-time across a number of computational platforms (unlike HOG or dense SIFT). Further computational speedups can be entertained if we start to skip frames to track. 

\begin{figure}[t] 
\includegraphics[width=0.32\textwidth]{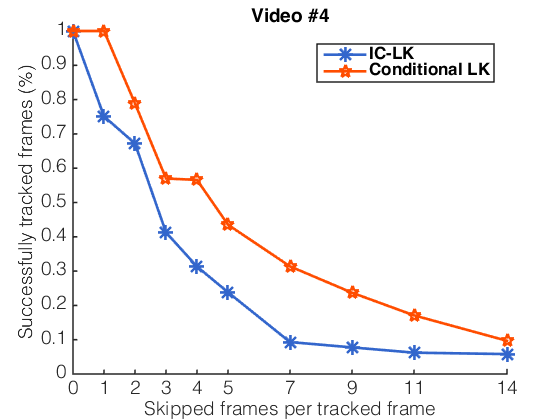} 
\includegraphics[width=0.32\textwidth]{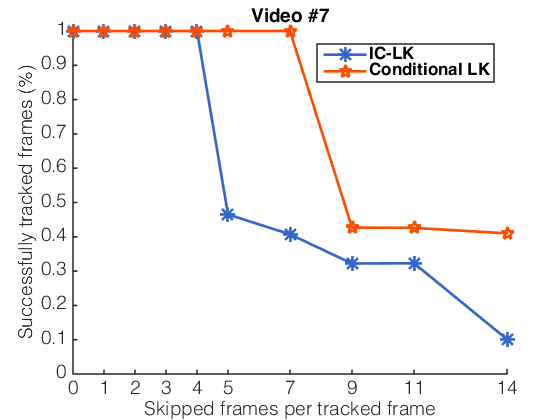} 
\includegraphics[width=0.32\textwidth]{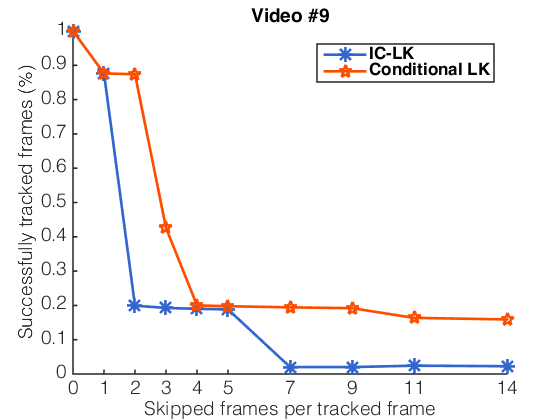} 
\centering
\caption{Tracking performance using IC-LK and Conditional LK with dense LBP descriptors for three videos under low frame-rate conditions, with and without lighting variations.}
\label{fig:bitplanes_plot}
\end{figure}

We compare the performance of Conditional LK with IC-LK and run the experiments on the videos collected in~\cite{DBLP:journals/corr/AlismailBL16}. We train the Conditional LK tracker on the first frame with $20$ synthetic examples. During tracking, we skip every $k$ frames to simulate low frame-rate videos. Figure \ref{fig:bitplanes_plot} illustrates the percentage of successfully tracked frames over the number of skipped frames $k$. It is clear that the Conditional LK tracker is more stable and tolerant to larger displacements between frames. 

\begin{figure}[t]
\includegraphics[width=0.24\textwidth]{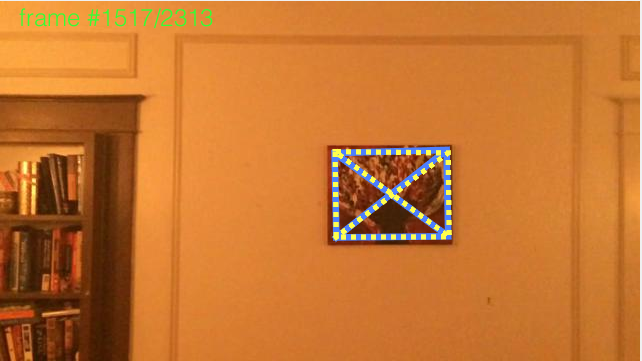} 
\includegraphics[width=0.24\textwidth]{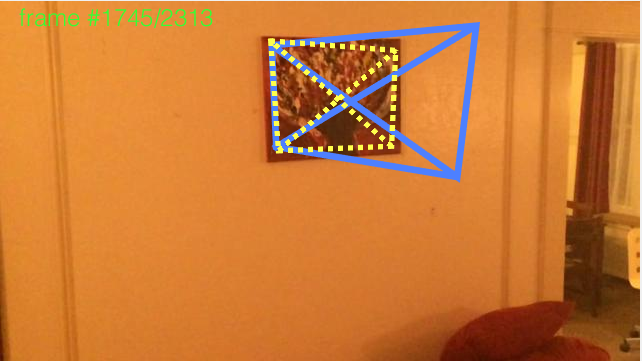} 
\includegraphics[width=0.24\textwidth]{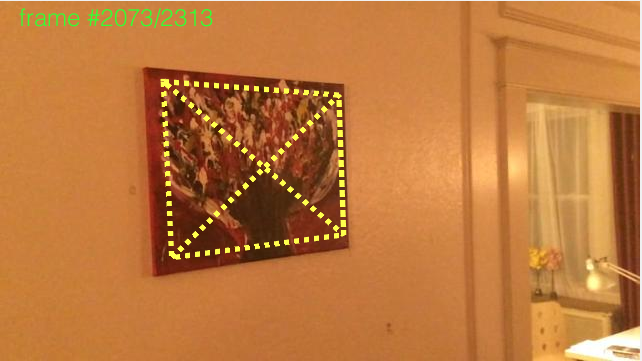} 
\includegraphics[width=0.24\textwidth]{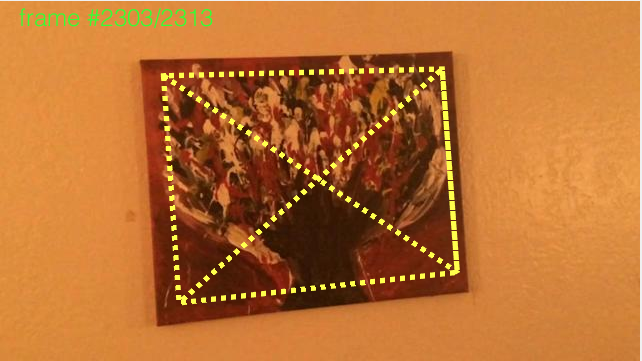} 
\includegraphics[width=0.24\textwidth]{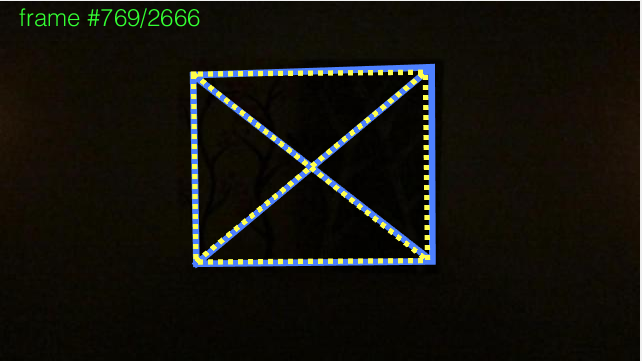} 
\includegraphics[width=0.24\textwidth]{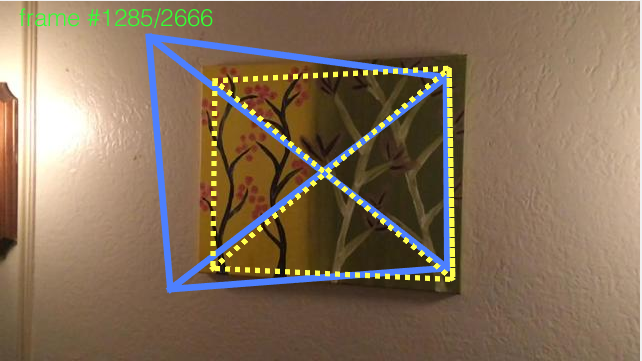} 
\includegraphics[width=0.24\textwidth]{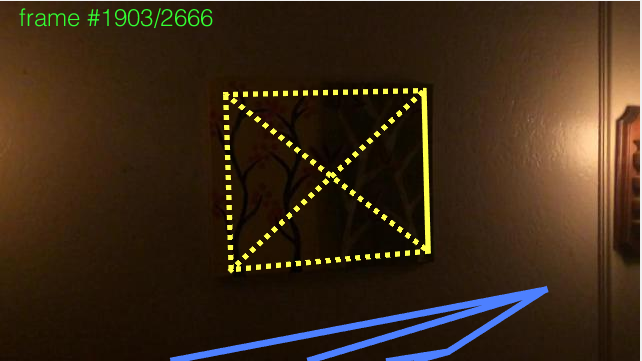} 
\includegraphics[width=0.24\textwidth]{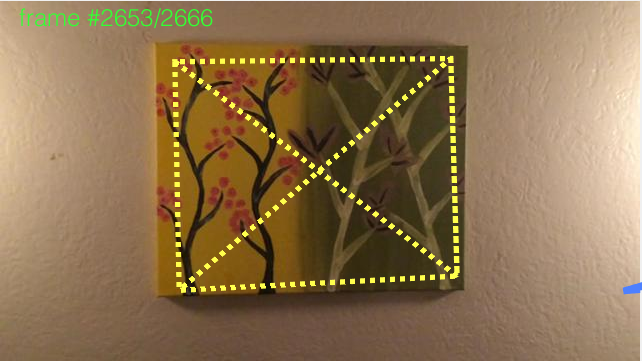} 
\includegraphics[width=0.24\textwidth]{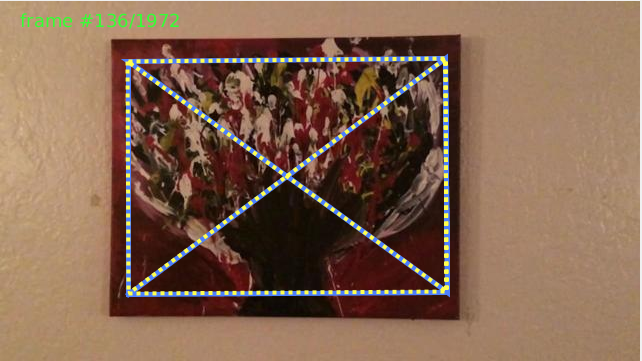} 
\includegraphics[width=0.24\textwidth]{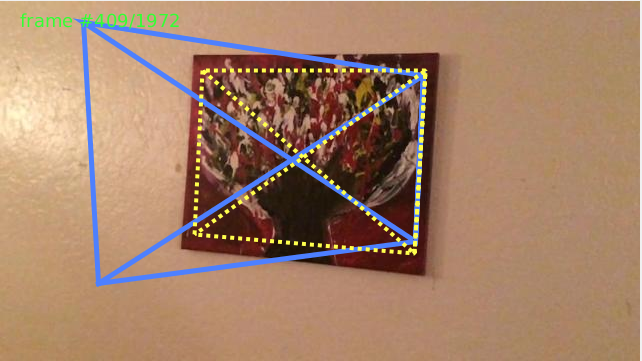} 
\includegraphics[width=0.24\textwidth]{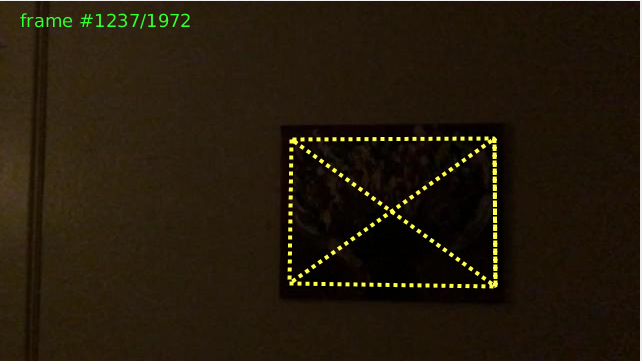} 
\includegraphics[width=0.24\textwidth]{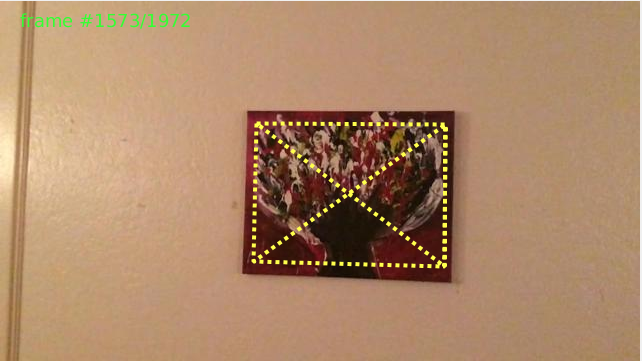} 
\centering
\caption{Snapshots of tracking results. Blue: IC-LK; yellow: Conditional LK. The second image of each row shows where IC-LK fails but Conditional LK still holds.}
\label{fig:bitplanes_v4s2}
\end{figure}

Figure \ref{fig:bitplanes_v4s2} shows some snapshots of the video, including the frames where the IC-LK tracker starts to fail but the Conditional LK tracker remains. This further demonstrates that the Conditional LK tracker maintains the same robustness to brightness variations by entertaining dense descriptors, but meanwhile improves upon convergence. Enhanced susceptibility to noises both in motion and brightness also suggests possible extensions to a wide variety of tracking applications.

\subsection{Facial Model Fitting}
In this experiment, we show how Conditional LK is applicable not only to 2D planar warps like affine or homography, but also to more complex warps that requires heavier parametrization. Specifically, we investigate the performance of our approach with a point distribution model (PDM)~\cite{matthews2004active} on the IJAGS dataset~\cite{matthews2004active}, which contains an assortment of videos with hand-labeled facial landmarks. We utilize a pretrained 2D PDM learned from all labeled data as the warp Jacobian and compare the Conditional LK approach against IC-LK (it has been shown that there is an IC formulation to facial model fitting~\cite{matthews2004active}). For Conditional LK, we learn a series of regressor/template pairs with $5$ examples per layer; for IC-LK, the template image is taken by the mean appearance.

\begin{figure}[t]
\includegraphics[width=\textwidth]{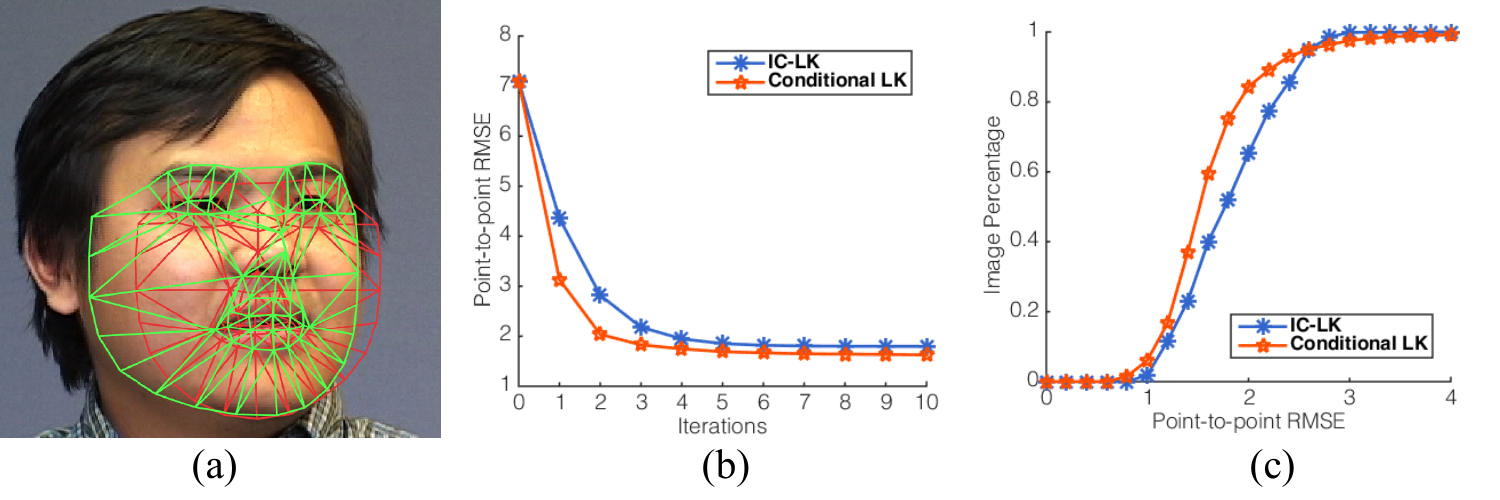} 
\centering
\caption{(a) An example of facial model fitting. The red shape indicates the initialization, and the green shape is the final fitting result. (b) Convergence rate comparison between IC-LK and Conditional LK. (c) Comparison of fitting accuracy.}
\label{fig:face}
\end{figure}

Figure \ref{fig:face} shows the results of fitting accuracy and convergence rate of subject-specific alignment measured in terms of the point-to-point RMSE of the facial landmarks; it is clear that Conditional LK outperforms IC-LK in convergence speed and fitting accuracy. This experiment highlights the possibility of extending our proposed Conditional LK to more sophisticated warps. We would like to note that it is possible to take advantage of the Conditional LK warp swapping property to incorporate a 3D PDM as to introduce 3D shape modelling; this is beyond the scope of discussion of this paper.

\section{Conclusion}
In this paper, we discuss the advantages and drawbacks of the LK algorithm in comparison to SDMs. We argue that by enforcing the pixel independence assumption into a conditional learning strategy we can devise a method that: (i) utilizes substantially less training examples, (ii) offers a principled strategy for regularization, and (iii) offers unique properties for adapting and modifying the warp function after learning. Experimental results demonstrate that the Conditional LK algorithm outperforms both the LK and SDM algorithms in terms of convergence. We also demonstrate that Conditional LK can be integrated with a variety of applications that potentially leads to other exciting avenues for investigation. 


\bibliographystyle{splncs}
\bibliography{reference}

\section*{Appendix I: Math Derivations of the Conditional LK Algorithm}

We describe the derivation and a few optimization details of the proposed Conditional LK algorithm. For convenience, we repeat the objective here,
\begin{eqnarray} \label{Eq:CLK}
\min_{\nabla \T(\0)} & \sum_{n \in \S} & \eucsqnorm{\Delta \p_{n} - \R[\I(\p_{n} \circ \Delta \p_{n} ) - \T(\0)]}  \\
 & \mbox {s.t. } & \R = \left( \begin{bmatrix} \nabla \T(\x_{1}) & \hdots & 0 \\ \vdots & \ddots & \vdots \\ 0 & \hdots & \nabla \T(\x_{D}) \end{bmatrix} 
\begin{bmatrix}
\frac{\partial \mathcal{W}(\x_{1}; \0)}{\partial \p^{\top} } \\
\vdots \\
\frac{\partial \mathcal{W}(\x_{D}; \0)}{\partial \p^{\top} } \nonumber
\end{bmatrix} \right)^{\dagger} ,
\end{eqnarray}
where
\begin{equation}
\nabla \T(\0) = \begin{bmatrix} \nabla \T (\x_{1}) \\ \vdots \\ \nabla \T(\x_{D}) \end{bmatrix} \nonumber
\end{equation}
is the compact form of the template ``image gradients'' we want to learn.
For simplicity, we further denote
$\g = \vec(\nabla \T(\0)) \in \mathbb{R}^{2KD}$ to be the vectorized form of $\nabla \T(\0)$, and we use $\Rg$ here instead of $\R$ to emphasize it is a function of $\g$.
Thus we can rewrite Equation~\ref{Eq:CLK} as
\begin{eqnarray}  \label{Eq:CLK2}
\min_{\g} & \sum_{n \in \S} & \eucsqnorm{\Delta \p_{n} - \Rg[\I(\p_{n} \circ \Delta \p_{n} ) - \T(\0)]}  \\
 & \mbox {s.t. } & \Rg = \left( \Gg \dxdp \right)^{\dagger} ,  \nonumber
\end{eqnarray}
where
\begin{equation}
\Gg = \mathcal{G}\left(\nabla \T(\0)\right) = \begin{bmatrix} \nabla \T(\x_{1}) & \hdots & 0 \\ \vdots & \ddots & \vdots \\ 0 & \hdots & \nabla \T(\x_{D}) \end{bmatrix} \;. \nonumber
\end{equation}

We can expand the pseudo-inverse form of $\Rg$ to be 
\begin{align} \label{eq:Rg}
\Rg = \left(\Hg\right)^{-1} \left(\dxdp\right)^{\top} \Gg^{\top} ,
\end{align}
where 
\begin{align*}
\Hg = \left(\dxdp\right)^{\top} \Gg^{\top} \Gg \dxdp
\end{align*}
is the pseudo-Hessian matrix. By the product rule, the derivative of $\Rg$ with respect to the $j$th element of $\g$, denoted as $\gj$, becomes
\begin{align} \label{eq:dRgdgj}
\dRgdgj = \dinvHgdgj \left(\dxdp\right)^{\top} \Gg^{\top} + \Hg^{-1} \left(\dxdp\right)^{\top} \indMtrxj^{\top} ,
\end{align}
where $\indMtrxj = \frac{\partial \Gg}{\partial \gj}$ is an indicator matrix with only the element in $\Gg$ corresponding to $\gj$ being active. The derivative of $\left(\Hg\right)^{-1}$ with respect to $\gj$ is readily given as
\begin{align} \label{eq:dHg-1dgj}
\dinvHgdgj = - \left(\Hg\right)^{-1} \dHgdgj \left(\Hg\right)^{-1} ,
\end{align}
where
\begin{align} \label{eq:dHgdgj}
\dHgdgj = \left(\dxdp\right)^{\top} \left( \Gg^{\top} \indMtrxj + \indMtrxj^{\top} \Gg \right) \dxdp \;.
\end{align}

Now that we have obtained explicit expression of $\dRgdg$, we can optimize $\g$ through gradient-based optimization methods by iteratively solving for $\deltag$, the updates to $\g$. One can choose to use first-order methods (batch/stochastic gradient descent) or second-order methods (Gauss-Newton or Levenberg-Marquardt). In the second-order method case, for examples, we can first rewrite Equation \ref{Eq:CLK2} in the vectorized form as
\begin{eqnarray} \label{eq:obj_condLK_vec}
\min_{\g} & \sum_{n \in \S} & \eucsqnorm{\Delta \p_{n} - \left[ (\I(\p_{n} \circ \Delta \p_{n} ) - \T(\0))^{\top} \kron \mathbf{I}_P \right] \vec(\Rg)} \;,
\end{eqnarray}
where $\mathbf{I}_P$ is the identity matrix of size $P$. Then the iterative update $\deltag$ is obtained by solving the least-squares problem
\begin{eqnarray*} \label{eq:obj_condLK_2nd}
\min_{\deltag} & \sum_{n \in \S} & \eucsqnorm{\Delta \p_{n} - \left[ (\I(\p_{n} \circ \Delta \p_{n} ) - \T(\0))^{\top} \kron \mathbf{I}_P \right] \vec( \mathcal{R}\left(\g + \deltag)\right)} \;,
\end{eqnarray*}
where $\vec( \mathcal{R}\left(\g + \deltag)\right)$ is linearized around $\g$ to be
\begin{eqnarray*} 
\vec( \mathcal{R}\left(\g + \deltag)\right) \approx \vec(\Rg) + \frac{\partial \vec(\Rg)}{\partial \g^{\top}} \deltag \;.
\end{eqnarray*}

Finally, the Conditional LK regressors $\R$ are formed to be
\begin{align} \label{eq:solve_condLK}
\R = \Rg = \left( \Gg \dxdp \right)^{\dagger} \;.
\end{align}

\end{document}